# Estimating the Potential Speedup of Computer Vision Applications on Embedded Multiprocessors

Vítor Schwambach*†, Sébastien Cleyet-Merle†, Alain Issard†, Stéphane Mancini*
* Univ. Grenoble Alpes, TIMA, F-38031 Grenoble, France
CNRS, TIMA, F-38031 Grenoble, France
† STMicroelectronincs, Grenoble, France

*Abstract*—Computer vision applications constitute one of the key drivers for embedded multicore architectures. Although the number of available cores is increasing in new architectures, designing an application to maximize the utilization of the platform is still a challenge. In this sense, parallel performance prediction tools can aid developers in understanding the characteristics of an application and finding the most adequate parallelization strategy. In this work, we present a method for early parallel performance estimation on embedded multiprocessors from sequential application traces. We describe its implementation in Parana, a fast trace-driven simulator targeting OpenMP applications on the STMicroelectronics' STxP70 Application-Specific Multiprocessor (ASMP). Results for the FAST key point detector application show an error margin of less than 10% compared to the reference cycle-approximate simulator, with lower modeling effort and up to 20x faster execution time.

## I. Introduction

In recent years, multiprocessor architectures have become ubiquitous in new embedded systems. Nonetheless, the development of efficient and scalable parallel applications, still represents a challenge[1]. Numerous factors can impact the parallel performance such as load imbalance, synchronization and communication overheads[3], that, if not accurately modeled, can lead to significant mismatches between simulation results and physical device measurements[8].

Different parallelization strategies can be used to optimize parallel performance. In order to evaluate them, developers often rely on time-consuming cycle-accurate simulations, or prototypes. Fast instruction set simulators or dynamic binary translation simulators trade some accuracy loss for faster simulation speeds. However, they all require a working parallel version of the application for each strategy, which can limit the exploration space due to the effort required to produce them in the first place. On the other hand, existing parallel performance prediction tools – Kismet[5], Parallel Prophet[6], and Intel Advisor XE[4] – focus solely on desktop-class applications and do not address embedded multiprocessors.

Our *main idea* is to acquire traces from a sequential application run on a reference simulator, then use a trace-driven simulator to estimate the parallel performance of the application on the target multiprocessor.

The *contributions* of this paper may be summarized as follows: (*i*) we propose a methodology for parallel application performance prediction from sequential code; (*ii*) we implement this methodology in Parana, and show that it can predict the parallel performance up to 20x faster than cycle-approximate simulators, with a margin of error in the order of 10% for the benchmarked application.

## II. Parallel Performance Estimation

Our methodology consists in designing an abstract trace-driven simulator to estimate the performance of an embedded application for a given parallelization strategy. Figure 1(a) depicts the four steps of the parallelization prediction and analysis flow using our Parana tool, which are detailed in the sequel.

**Step 1: Platform Characterization.** The first step is to characterize the target multiprocessor platform and its parallel programming framework in order to build a characterization database. This database contains statistical information of the measured overheads for the OpenMP directives, as well as inherent characteristics of the multiprocessor platform, such as memory latency and bandwidth parameters for each memory hierarchy level. For this, an enriched version of the EPCC OpenMP micro-benchmarks[2] is used. The traces generated by a reference simulator or prototype are fed to a characterization tool that generates this database. The characterization database only needs to be generated once for a given multiprocessor platform and can be reused in later steps.

**Step 2: Application Trace Collection.** In this phase we collect execution traces from the sequential application. Tasks are created from function calls in the application, which are annotated with their timestamps and hierarchical information. Summary memory access statistics are also gathered for each task. This task trace can be enriched by user inserted instrumentation macros that (*i*) define explicit tasks at lower granularity levels and (*ii*) add additional semantic information, such as if the tasks refer to loops, critical regions or array accesses, among others.

**Step 3: Parallelization Directives Specification.** This step consists in creating a file with the OpenMP parallelization directives the user wishes to evaluate, and a mapping of these directives onto functions or user-defined code sections.

**Step 4: Parallelization Analysis.** In this final step, Parana uses the platform characterization and application trace databases, as well as the parallelization directives, to predict the parallel performance of the application. It first builds a directed acyclic graph of the application's tasks and generates multiple parallel schedules, according to the parallelization directives. It acquires statistics from each parallel schedule to produce a detailed parallelization report. The user can then refine the application and repeat the process to explore new parallelization strategies.

## III. Results

### A. Target Platform

Our target platform is the STxP70 ASMP. This multiprocessor is similar to a single cluster of the STHORM platform[7]. It has a configurable SMP architecture with up to 16 STxP70 cores – 32-bit dual-issue RISC CPUs. Heterogeneity can be achieved by having multiple instances with particular configurations and instruction set extensions or by attaching dedicated HW IPs to DMA queues. The STxP70 ASMP architectural template is depicted in Figure 1(b).

### B. Execution Vehicles

**Gepop Simulator.** Gepop is a cycle-approximate simulator for the STxP70 ASMP platform and constitutes our reference platform for the platform characterization.

**FPGA Prototype.** A prototype of the STxP70 ASMP on the Xilinx VC707 FPGA is used for comparison.

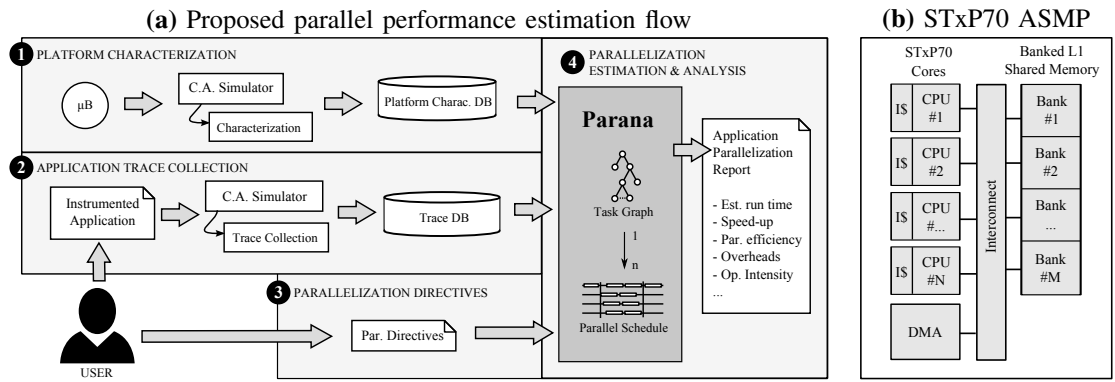

Fig. 1: **(a)** Overview of the four steps of the proposed flow for parallel application performance prediction with Parana. **(b)** Architectural template of the STxP70 ASMP.

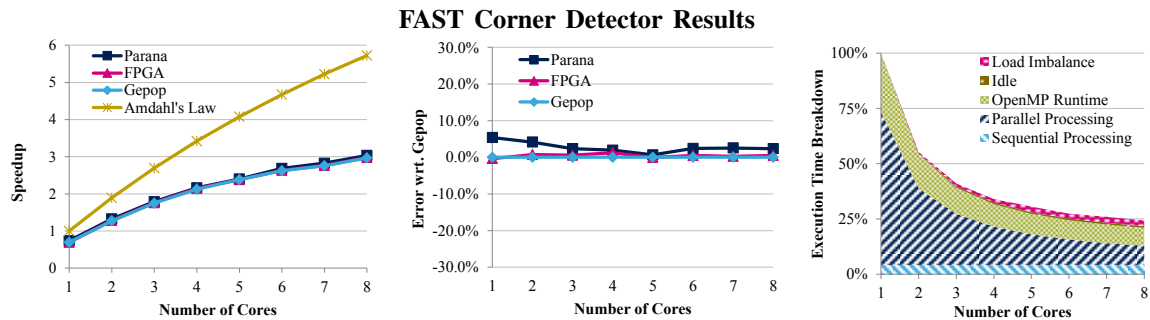

Fig. 2: Comparison of speedup estimates with Parana and measurements with the Gepop cycle-approximate simulator (Parana's characterization reference) and an FPGA prototype for the FAST Corner Detection application. Left: speedup results from 1 to 8 cores. Center: speedup error wrt. Gepop. Right: Cycle stacks showing the impact of different overhead sources.

**Parana.** The proposed tool is a trace-diven simulator that can estimate the parallel performance of an application from its sequential execution traces and a set of OpenMP directives.

*C. Application*

**FAST Corner Detection (FAST).** We have ported a 9-16 FAST corner detector (from OpenCV 2.4.6) commonly used in computer vision to select key points in a number of tracking and detection applications.

*D. Results*

Figure 2 provides the speedup results, as well as the relative error with respect to the reference Gepop simulator, for the FAST application. Parana's speedup estimates have an average mean percentage error of 3% – with a maximum absolute error of 5.8%. Parana's cycle stacks show that the scalability bottleneck is the high OpenMP runtime overhead. This knowledge allows the user to adopt strategies to decrease the scheduling time, such as opting for a higher parallelization granularity.

As for the execution time, Parana was on average 15.9x faster than the Gepop simulator – with some cases presenting up to 20x faster execution time. The simulation with the FPGA prototype was slower due to the time it takes to load the application code and due to interactions with the host PC in system calls, resulting in a 36.7x faster execution time with Parana.

## IV. CONCLUSION

This paper presents a methodology and a tool – Parana – for early parallel performance estimation from sequential application traces. The results of the proposed methodology were compared against metrics obtained from the execution of an OpenMP version of the application on the Gepop cycle-approximate simulator and on an FPGA prototype of the STxP70 ASMP. We have demonstrated the accuracy of the solution in estimating the parallel speedup, within 10% of the reference simulator, as well as its interest in identifying the sources of scalability issues via the cycle stacks. Future explorations will aim at improving design space exploration capabilities, as well as modeling DMA transfers and shared memory conflicts to increase the domain of addressable applications.

## REFERENCES

[1] K. Asanovic, R. Bodik, J. Demmel, T. Keaveny, K. Keutzer, J. Kubiatowicz, N. Morgan, D. Patterson, K. Sen, J. Wawrzynek, and Others. A view of the parallel computing landscape. *Communications of the ACM*, 52(10):56–67, 2009.
[2] J. M. Bull and D. O'Neill. A microbenchmark suite for OpenMP 2.0. *ACM SIGARCH Computer Architecture News*, 29(5):41–48, 2001.
[3] W. Heirman, T. E. Carlson, S. Che, K. Skadron, and L. Eeckhout. Using cycle stacks to understand scaling bottlenecks in multi-threaded workloads. In *Workload Characterization (IISWC), 2011 IEEE International Symposium on*, pages 38–49. IEEE, Ieee, Nov. 2011.
[4] Intel Corporation. Intel Advisor XE. https://software.intel.com/en-us/intel-advisor-xe, last accessed in November 2014.
[5] D. Jeon, S. Garcia, C. Louie, and M. Taylor. Kismet: parallel speedup estimates for serial programs. *ACM SIGPLAN Notices*, 2011.
[6] M. Kim, P. Kumar, H. Kim, and B. Brett. Predicting Potential Speedup of Serial Code via Lightweight Profiling and Emulations with Memory Performance Model. *2012 IEEE 26th International Parallel and Distributed Processing Symposium*, pages 1318–1329, May 2012.
[7] J. Mottin, M. Cartron, and G. Urlini. The STHORM Platform. In *Smart Multicore Embedded Systems*, pages 35–43. Springer, 2014.
[8] V. Schwambach, S. Cleyet-Merle, A. Issard, and S. Mancini. Application-level Performance Optimization: A Computer Vision Case Study on STHORM. *Procedia Computer Science*, 29:1113–1122, 2014.